\documentclass[letterpaper]{article} 
\usepackage{aaai25}  
\usepackage{times}  
\usepackage{helvet}  
\usepackage{courier}  
\usepackage[hyphens]{url}  
\usepackage{graphicx} 
\usepackage{caption} 
\usepackage{amsthm} 
\usepackage{amsfonts}
\usepackage{color}
\usepackage{amsmath}
\usepackage{amssymb}
\usepackage{soul}
\usepackage{amsthm}
\usepackage[ruled,vlined]{algorithm2e}
\newtheorem{definition}{Definition}
\newtheorem{remark}{Remark}

\nocopyright

\usepackage{listings}
\DeclareCaptionStyle{ruled}{labelfont=normalfont,labelsep=colon,strut=off} 
\title{Automaton Distillation:  Neuro-Symbolic Transfer Learning for Deep
Reinforcement Learning}
\author{
    Suraj Singireddy\textsuperscript{\rm 1}, 
    Precious Nwaorgu\textsuperscript{\rm 2}, 
    Andre Beckus\textsuperscript{\rm 3}, 
    Aden McKinney\textsuperscript{\rm 2}, 
    Chinwendu Enyioha\textsuperscript{\rm 2}, 
    Sumit Kumar Jha\textsuperscript{\rm 4}, 
    George K. Atia\textsuperscript{\rm 2}, 
    Alvaro Velasquez\textsuperscript{\rm 5}
}
\affiliations{
    \textsuperscript{\rm 1}University of Texas at San Antonio\\
     \textsuperscript{\rm 2}University of Central Florida\\
     \textsuperscript{\rm 3}Air Force Research Laboratory\\
       \textsuperscript{\rm 4}Florida International University\\
        \textsuperscript{\rm 5}University of Colorado\\

}

\usepackage{bibentry}

\begin{document}

\maketitle

\begin{abstract}
Reinforcement learning (RL) is a powerful tool for finding
optimal policies in sequential decision processes. However, deep RL methods have two weaknesses: collecting the
amount of agent experience required for practical RL problems is prohibitively expensive, and the learned policies exhibit poor generalization on tasks outside the training data distribution. To mitigate these issues, we introduce automaton
distillation, a form of neuro-symbolic transfer learning in
which Q-value estimates from a teacher are distilled into a
low-dimensional representation in the form of an automaton. We then propose methods for generating Q-value estimates
where symbolic information is extracted
from a teacher's Deep Q-Network (DQN). The resulting Q-value estimates are used to bootstrap
learning in the target discrete and continuous environment via a modified DQN and Twin-Delayed Deep Deterministic (TD3) loss function, respectively. We demonstrate that automaton distillation decreases the time required to
find optimal policies for various decision tasks in new environments, even in a target environment different in structure from the source environment.
\end{abstract}

\section{Introduction}
Sequential decision tasks, in which an agent seeks to learn
a policy to maximize long-term reward through trial and error, are often solved using reinforcement learning (RL) approaches. These approaches must balance exploration (the
acquisition of novel experiences) with exploitation (taking the predicted best action based on previously acquired
knowledge). Performing sufficient exploration to find the optimal policy requires collecting much experience, which can be expensive.

One explanation for the high sample efficiency observed in human learning relative to neural networks is the ability to apply high-level concepts learned through prior experience
to environments not encountered during training. Although
traditional deep learning methods do not contain an explicit
notion of abstraction, neuro-symbolic computing has shown promise as a way to integrate high-level symbolic reasoning into neural approaches. 
Symbolic logic provides a formal mechanism for injecting and extracting knowledge from
neural networks to guide learning and provide explainability,
respectively \cite{tran2016deep}.  Additionally, RL over
policies expressed as symbolic programs has been shown to improve performance and generalization on previously unseen tasks ~\cite{verma2018programmatically,anderson2020neurosymbolic}.

In this paper, we adopt a different approach which uses a symbolic representation of RL objectives to facilitate knowledge transfer between an expert in a related source domain (the `teacher') and an agent learning the target task (the `student'). Although it is common to use reward signals to convey an objective, many decision tasks can be more naturally expressed as a high-level description of the intermediate steps required to achieve the objective in the form of natural language. These natural language descriptions can be translated into a corresponding specification in a formal language such as linear temporal logic (LTL)~\cite{brunello2019synthesis}, which can be converted into an equivalent automaton representation~\cite{wolper1983reasoning}. For tasks which share an objective, the automaton acts as a common language; states and actions in the source and target domains can be mapped to nodes and transitions in the automaton, respectively. Furthermore, by assigning value estimates to automaton transitions, the automaton representation can be transformed into a compact model of the environment to facilitate learning.

We introduce two new variants of transfer learning that, to the best of our knowledge, have not been previously explored in the literature. These variants leverage the \emph{automaton representation of an objective} to convey information about the reward signal from the teacher to the student. The first, static transfer, generates estimates of the Q-value of automaton transitions by performing value iteration over the abstract Markov Decision Process (MDP) defined by the automaton. The second variant, dynamic transfer, distills knowledge from a teacher Deep Q-Network (DQN) into the automaton by mapping teacher Q-values of state-action pairs in the experience replay buffer to their corresponding transition in the automaton. 

\begin{figure}[t]
    \centering
    \includegraphics[width=7cm]{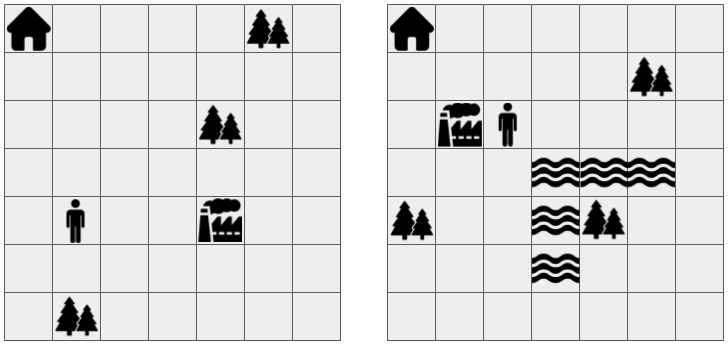} \\
     (a) \hspace{26mm} (b)
     
     \includegraphics[width=7cm]{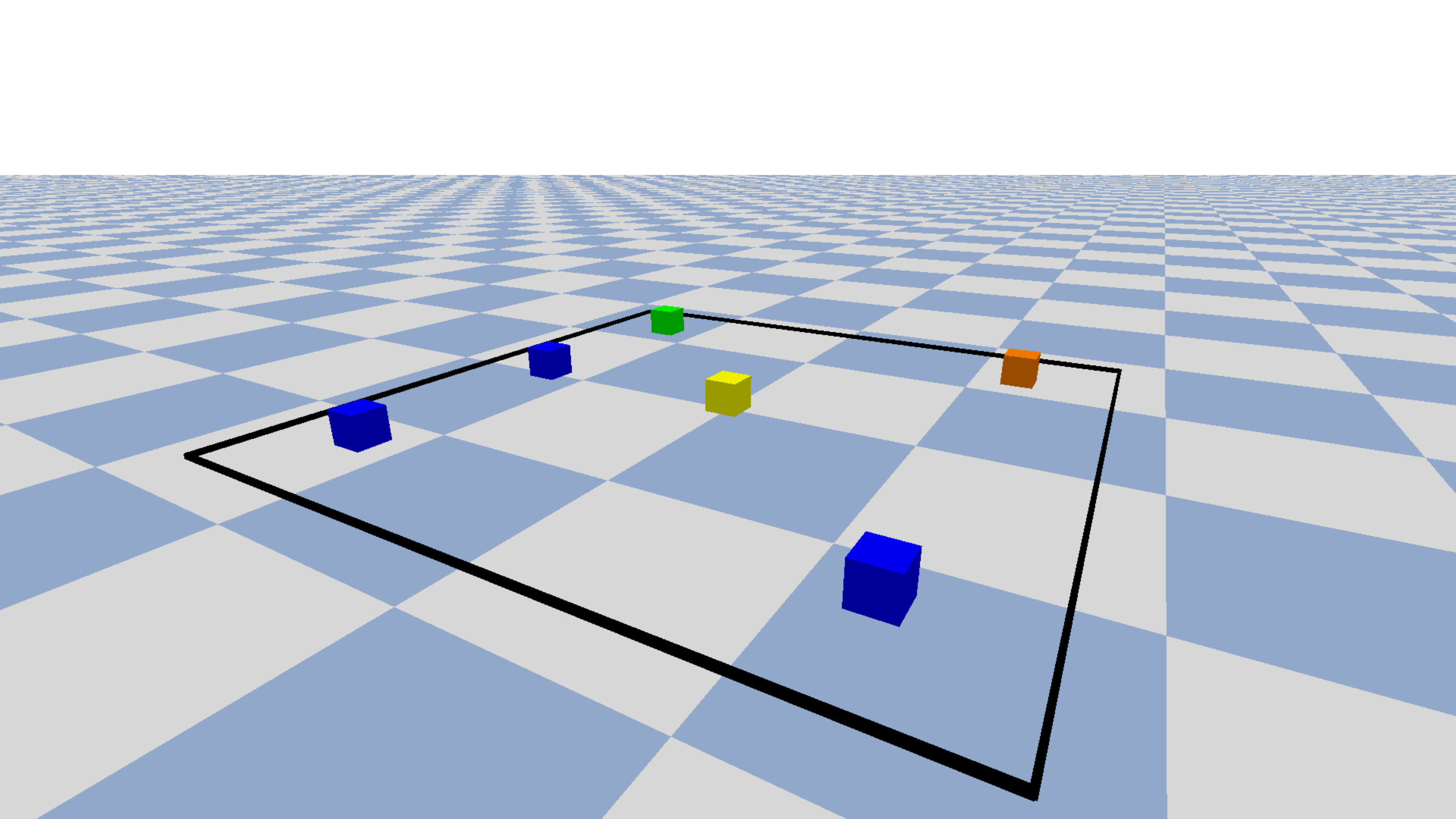}\\
     (c)
    \caption{Example environment configurations for the Blind Craftsman teacher (a) and student (b) environments with additional obstacles introduced. (c) A continuous state and action student environment where the yellow, green, and orange cubes represent the agent, wood, factory, and home, respectively, placed at random positions in the continuous space.}
    \label{fig:blind_craftsman_obstacles}
\end{figure}

Q-value estimates generated through either of these methods can be used to bootstrap the student's learning process. We evaluate our method on scenarios (such as that presented in Figure~\ref{fig:blind_craftsman_obstacles}) where: \textit{i)} the student environment has features absent in the teacher's task; and \textit{ii)} the student operates within a different state and action space, that is transferring to a student in a continuous environment. These scenarios demonstrate the robustness of the transferred knowledge to varying environmental conditions. Additionally, we argue that certain other automaton-based transfer methods might induce \textit{negative transfer} in some scenarios. We demonstrate the efficacy of our proposed method in reducing training time, even when existing methods harm performance.

In the next section, we introduce the building blocks for the problem including Non-Markovian Reward Decision Processes (NMRDP) as the model for the sequential decision problem with non-Markovian reward signal, the Deterministic Finite Automata (DFA) used to represent a reduced order model for the solution to the sequential decision problem, and the Cross-Product MDP, which allows for the use of RL algorithms on the NMRDP. After that, we present the knowledge transfer algorithm using the Automaton as the distillation vehicle. Following that, we place our work in context by comparing it to some existing approaches in transfer learning and follow with experiments. In the experiments, we illustrate performance of the proposed method on transferring knowledge between similarly structured (discrete-to-discrete environments) and structurally different (discrete-to-continuous) environment.

\section{Preliminaries}
\label{sec:prelim}
We assume that the teacher and student decision processes are non-Markovian Reward Decision Processes (NMRDP). 

\vspace{3pt}
\begin{definition}[Non-Markovian Reward Decision Process (NMRDP)]
    An NMRDP is a decision process defined by the tuple $\mathcal{M} = \langle S, s_0, A, T, R \rangle$, where $S$ is the set of valid states, $s_0 \in S$ is the initial state, $A$ is the set of valid actions, $T: S \times A \times S \rightarrow [0,1]$ is a transition function defining transition probabilities for each state-action pair to every state in $S$, and $R: (S \times A)^* \rightarrow \mathbb{R}$ defines the reward signal observed at each time step based on the sequence of previously visited states and actions.
\end{definition}

NMRDPs differ from MDPs in that the reward signal $R$ may depend on the entire history of observations rather than only the current state. However, the reward signal is often a function of a set of abstract properties of the current state, which is of much smaller dimension than the original state space. Thus, it can be beneficial to represent the reward signal in terms of a simpler vocabulary defined over features extracted from the state. We assume the existence of a set of atomic propositions $AP$ for each environment, which capture the dynamics of the reward function, as well as a labeling function, $L: S \rightarrow 2^{AP}$, that translates experiences into truth assignments for each proposition $p \in AP$. In cases where such a labeling function does not explicitly exist, it is possible to find one automatically~\cite{hasanbeig2021deepsynth}. Furthermore, we assume that the objectives in the teacher and student environments are identical and represented by the automaton $\mathcal{A} = (2^{AP}, \Omega, \omega_0, F, \delta)$ as defined below.

\begin{figure}[t]
	\centering
	$\begin{matrix}

	\includegraphics[width=8cm]{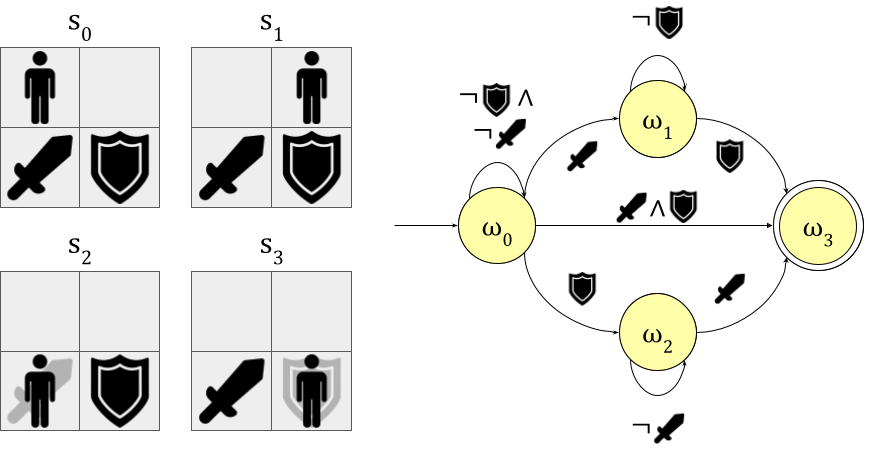}
	\end{matrix}$ \\
	\hspace{-4mm} (a) \hspace{38mm} (b)
    \caption{
    (a) A simple NMRDP. At each time step, the agent may move one square in any cardinal direction. A sequence of actions satisfies the objective if and only if the agent obtains both the sword and the shield. The objective is decomposed using the atomic propositions $AP = \{\text{sword}, \text{shield}\}$, with a labeling function $L$ such that $L(s_0) = \{\}, L(s_1) = \{\}, L(s_2) = \{\text{sword}\}, L(s_3) = \{\text{shield}\}$. Rollouts which achieve the objective also satisfy the LTL$_f$ specification $\phi = \textbf{F}(\text{sword}) \wedge \textbf{F}(\text{shield})$.
    (b) An automaton defined over the alphabet $\Sigma = \{\{\}, \{\text{sword}\}, \{\text{shield}\}, \{\text{sword, shield}\}\}$. The automaton accepts the subset of strings in $\Sigma^*$ that satisfy the LTL$_f$ formula. 
    }
	\label{fig:simplest}
\end{figure}

\vspace{3pt}
\begin{definition}[Deterministic Finite-State Automaton (DFA)]
   
    A DFA is an automaton defined by the tuple $\mathcal{A} = \langle \Sigma, \Omega, \omega_0, F, \delta \rangle$, where $\Sigma$ is the alphabet of the input language, $\Omega$ is the set of states with starting state $\omega_0$, $F \subseteq \Omega$ is the set of accepting states, and $\delta: \Omega \times \Sigma \rightarrow \Omega$ defines a state transition function.
	\label{Buchi}
\end{definition}

The atomic propositions $AP$ comprise a vocabulary of abstract properties of the state space which directly correspond to the reward structure. Using the labeling function $L$, states in an NMRDP can be mapped to an element in the alphabet $\Sigma = 2^{AP}$. Then, the set of rollouts which satisfy the objective constitute a regular language over $\Sigma$.
The parameters $\Omega, \omega_0, F, \delta$ are chosen such that the set of strings accepted by the objective automaton $\mathcal{A}$ is equivalent to the aforementioned regular language; we illustrate this with a simple example in Figure~\ref{fig:simplest}. An additional consequence of developing such a vocabulary is that RL objectives can be expressed as a regular language and subsequently converted into a DFA.
\begin{remark}
One benefit of the automaton representation is the ability to express non-Markovian reward signals in terms of the automaton state. During an episode, the state of the automaton can be computed in parallel with observations from the learning environment.     
\end{remark}
Given the current automaton state $\omega$ and a new observation $s'$, the new automaton state can be computed as $\omega' = \delta(\omega, L(s'))$. We assume that the atomic propositions capture the non-Markovian behavior of the reward signal, thus, the system dynamics are Markovian in the cross-product of the observation and automaton state spaces. Formally, we represent the cross-product of an NMRDP and its corresponding objective automaton as an MDP.

\vspace{3pt}
\begin{definition}[Cross-Product Markov Decision Process]
    The cross-product of an NMRDP $\mathcal{M} = \langle S, s_0, A, T, R \rangle$ and a DFA $\mathcal{A} = \langle \Sigma, \Omega, \omega_0, F, \delta  \rangle$ which captures the non-Markovian behavior of the reward signal is a Markov Decision Process (MDP) $\mathcal{M}_\text{prod} = \langle S \times \Omega, (s_0,\omega_0), A \times \Sigma, T \times \delta, R' \rangle$ where $R' : \Omega \times \Sigma \rightarrow \mathbb{R}$ is a Markovian reward signal (i.e., can be expressed as a function of only the current state and action).
\end{definition}

Note that transforming an NMRDP into a cross-product MDP permits the use of traditional RL algorithms, which rely upon the Markovian assumption, for decision tasks with non-Markovian reward signals.

Implicit in the previous discussion is that the objective is translated into an automaton prior to learning. The difficulty of explicitly constructing an automaton to represent a desired objective has motivated the development of automated methods for converting a reward specification into an automaton representation. Such methods observe a correspondence between formal logics (such as regular expressions, LTL, and its variants) and finite-state automata~\cite{wolper1983reasoning}. In particular, finite-trace Linear Temporal Logic (LTL$_f$) has been used to represent RL objectives~\cite{camacho2019ltl,velasquez2021dynamic}.

\vspace{3pt}
\begin{definition}[Finite-Trace Linear Temporal Logic (LTL$_f$)]
	A formula in LTL$_f$ consists of a set of atomic propositions $AP$ which are combined by the standard propositional operators and the following temporal operators: the next operator \textbf{X}$\phi$ ($\phi$ will be true in the next time step), the eventually operator \textbf{F}$\phi$ ($\phi$ will be true in some future time step), the always operator \textbf{G}$\phi$ ($\phi$ will be true in all future time steps), the until operator $\phi_1$ \textbf{U} $\phi_2$ ($\phi_2$ will be true in some future time step, and until then $\phi_1$ must be true), and the release operator $\phi_1$ \textbf{R} $\phi_2$ ($\phi_2$ must be true always or until $\phi_1$ first becomes true).
\end{definition}

It has been shown that specifications in LTL$_f$ can be transformed into an equivalent deterministic B\"uchi automaton~\cite{de2015synthesis}, and tools for compiling automata are readily available~\cite{zhu2017symbolic}. Moreover, it is possible, in principle, to convert descriptions using a predefined subset of natural language into LTL~\cite{brunello2019synthesis}. Thus, it is feasible to translate a specification provided by a domain expert into an automaton representation using automated methods.

\section{Automaton Distillation}
\label{sec:autdistill}

This section describes our approach for transferring knowledge from a teacher to a student, which leverages a DQN expert trained in the teacher environment and an automaton representation of the objective. Knowledge is first distilled from the teacher into the automaton and then from the automaton into the student during training.
We term this methodology ``automaton distillation'', as it entails distilling insights from a teacher automaton to enhance the learning in the student.

Automaton distillation begins by training a teacher agent using Deep Q-learning. Subsequently, the teacher's Q-values are distilled into the objective automaton; that is, the automaton representing the sequence of tasks, associating each transition's value with an estimated Q-value for corresponding state-action pairs in the teacher environment. Then, a student agent undergoes training in the target environment, employing an adapted loss function that incorporates the Q-values derived from the automaton.

\noindent\textbf{Dynamic Transfer.} Our primary focus lies in a dynamic transfer learning algorithm that distills value estimates from an agent trained with Deep Q-learning into the objective automaton. This process uses a contractive mapping from the teacher NMRDP's state-action pairs to the abstract MDP defined by the automaton. Additionally, an analogous expansive mapping from the abstract MDP to the student NMRDP provides an initial Q-value estimate for state-action pairs in the target domain, assisting the student's learning process. The automaton mediates between the teacher and student domains, enabling experience sharing without requiring manual state-space mappings \cite{taylor2005behavior} or unsupervised map learning \cite{ammar2015autonomous}.

The \emph{teacher} DQN is trained using only standard RL methods~\cite{wang2016dueling,van2016deep,schaul2015prioritized}. However, to track both the current node in the automaton and the environment state, we store samples of the form $((s, \omega), a, r, (s', \omega'))$ in the experience replay $ER$ buffer. We define $\eta_\text{teacher}: \Omega \times \Sigma \to \mathbb{N}$ as the number of times each automaton node $\omega$ and a set of atomic propositions $\sigma \in 2^{AP}$ appear in the experience replay $ER$ of the teacher (note that $\omega$ and $\sigma$ define a transition in the automaton objective as given by $\delta(\omega, \sigma) = \omega'$):
\begin{equation}
    \eta_\text{teacher}(\omega, \sigma) = |\{((s, \omega), a, r, (s', \omega')) \in ER | L(s') = \sigma\}|.
\end{equation}
Similarly, we define $Q^\text{avg}_\text{teacher}: \Omega \times \Sigma \rightarrow \mathbb{R}$ to be the average Q-value corresponding to the automaton transition given by $\omega$ and $\sigma \in 2^{AP}$, according to the teacher DQN: 
\begin{equation}
   Q^\text{avg}_\text{teacher} (\omega, \sigma) \hspace{-0.65mm}= \hspace{-0.65mm}  \frac{\sum_{\{((s, \omega), a, r, (s', \omega')) \in ER | L(s') = \sigma\}}\hspace{-1mm} Q_\text{teacher} (\hspace{-0.4mm} s, \hspace{-0.3mm}a\hspace{-0.3mm})}{\eta_\text{teacher}(\omega, \sigma)}.
   \label{eq:q_teach_avg}
\end{equation}

In DQN and actor-critic networks, target networks are critical for stabilizing training by providing fixed targets for temporal difference (TD) updates. These networks are typically updated at a slower rate or through a smoothing process to avoid instability caused by rapidly changing Q-values.

In this paper, we incorporate the teacher's knowledge into the student's learning by modifying the \emph{student's target} Q-value:
\begin{multline}
    Q'_\text{student}(s, a) = \beta(\omega, L(s')) \, Q^\text{avg}_\text{teacher}(\omega, L(s')) \\
    + \big(1 - \beta(\omega, L(s'))\big) Q_\text{target}.
\label{equation:modified_target}
\end{multline}
Here, $\beta: \Omega \times \Sigma \rightarrow [0, 1]$ controls the influence of the teacher's knowledge, $Q^\text{avg}_\text{teacher}(\omega, L(s'))$ is the teacher's average Q-value for transition $(\omega, L(s'))$, and $Q_\text{target}$ is the standard target Q-value. This approach applies to both DQN and actor-critic algorithms like Deep Deterministic Policy Gradient (DDPG) \cite{lillicrap2015continuous} and Twin Delayed DDPG (TD3) \cite{fujimoto2018addressing}, where $Q_\text{target}$ in \eqref{equation:modified_target} is defined as follows. For DQN,
\begin{equation}
    Q_\text{target} = r + \gamma \max_{a'} Q_\text{student}(s', a'; \theta^\text{target}),
    \label{equation:DQN_target}
\end{equation}
and for DDPG and TD3,
\begin{equation}
    Q_\text{target} = r + \gamma Q_\text{student}\big(s', \pi_{\phi'}(s'); \theta^\text{target}\big).
    \label{equation:TD3_target}
\end{equation}
In these equations, $r$ is the immediate reward, $\gamma$ is the discount factor, $s'$ is the next state, $Q_\text{student}$ is the student's Q-value function, $\theta^\text{target}$ represents the parameters of the target network, and $\pi_{\phi'}(s')$ is the action suggested by the target policy network at state $s'$.

The annealing function is defined as $\beta(\omega, \sigma) = \rho^{\eta_\text{student}(\omega, \sigma)}$, where $\rho = 0.999$ and $\eta_\text{student}(\omega, \sigma)$ is the number of times the transition $(\omega, \sigma)$ has been sampled during student training.

The loss function for updating the student's network is:
\begin{equation}
    \begin{aligned}
        \mathrm{Loss}(\theta) = \mathbb{E}_{((s, \omega), a, r, (s', \omega')) \sim P(ER)}  [ \:
        &Q'_\text{student}(s,a)  \\
        &- Q(s, a; \theta)]^2,
        \label{equation:newLoss}
    \end{aligned}
\end{equation}
where $Q(s, a; \theta)$ is the student's Q-value prediction, and $P$ is a priority function favoring samples with higher prediction errors. 

By incorporating the teacher's Q-values from Equation \eqref{eq:q_teach_avg} into the student's target calculation in Equation \eqref{equation:modified_target}, especially early in training, we mitigate poor initial value estimates that slow convergence in standard RL algorithms. This method performs non-Markovian knowledge transfer, as the automaton compactly represents environment dynamics and encodes the non-Markovian reward signal. This versatility enables our method to facilitate knowledge transfer not only between discrete environments but also from discrete to continuous environments, as demonstrated in our experiments. Our approach enhances the student's learning by integrating the teacher's knowledge directly into the target Q-value. This allows the student to benefit from the teacher's expertise while adapting to its own environment, leading to faster convergence and improved performance in complex tasks.

\begin{algorithm}[tb]
    \caption{Automaton Distillation with DQN or TD3}
    \label{alg:algorithm_automaton}
    \SetAlgoLined
    
    Initialize Q-Networks: for {DQN:} $Q_{\theta_D}$; {TD3:} critic $Q_{\theta_1}, Q_{\theta_2}$; and TD3 actor $\pi_{\phi}$.

    Initialize target networks:
    \begin{itemize}
    \renewcommand\labelitemi{--}
    \item {DQN:} $\theta'_D \gets \theta_D$.
    \item {TD3:} $\theta'_1 \gets \theta_1$, $\theta'_2 \gets \theta_2$,$\phi' \gets \phi$.
  \end{itemize}

    Initialize replay buffer $ER$, automaton transition visit counts $\eta$\
    
    \For{$t \leftarrow 1$ \KwTo $T$}{
        Take action $a$ w.r.t. to $\pi_{\phi}$ or $Q_{\theta_D}$ , observing reward $r$ and new state $s'$\;
        Compute new automaton state $\omega' = \delta(\omega, L(s'))$\;
        Append augmented experience $((s,\omega), a, r, (s',\omega'))$ to the replay buffer $ER$ with priority 1\;
        Sample $M$ transitions from replay with priority $p_i$\;
        Compute annealing parameters for each transition $\beta_i \leftarrow \rho^{\eta(\omega_i,L(s'_i)}$\;
        update $Q_\text{target}$ from Eq. \eqref{equation:DQN_target} or Eq. \eqref{equation:TD3_target}
        
        Generate adjusted targets $Q'_i \leftarrow \beta_i Q_\text{teacher}(\omega_i, L(s'_i)) + (1 - \beta_i)Q_\text{target}$;
        
        Update Q-networks $\theta \leftarrow \theta + \frac{\alpha}{M} \sum_i p_i \nabla_\theta (Q'_i - Q((s_i,\omega_i),a_i;\theta))^2$\;
        Update buffer priorities $p_i = (Q'_i - Q((s_i,\omega_i),a_i;\theta))^2$\;
        \For{$i \leftarrow 1$ \KwTo $M$}{
            Update visit count $\eta(\omega_i,L(s'_i)) \leftarrow \eta(\omega_i,L(s'_i)) + 1$\;
        }
        \If{$t = 0 \mod \tau$}{
            $\theta_\text{D} \leftarrow \theta$\;

            $\nabla_{\phi} J(\phi) = M^{-1} \sum \nabla_a Q_{\theta_1}(s, a) \rvert_{a=\pi_{\phi}(s)} \nabla_{\phi} \pi_{\phi}(s);$
            $\theta'_i \gets \tau \theta_i + (1 - \tau) \theta'_i;$
            
            $\phi' \gets \tau \phi + (1 - \tau) \phi';$
        }
    }
\end{algorithm}

The asymptotic behavior of automaton Q-learning depends on the annealing function $\beta$ in the student Q-udpate in \eqref{equation:modified_target}. When $\beta = 0$, automaton Q-learning reduces to vanilla Q-Learning. 
Next, we establish a convergence result for tabular automaton Q-learning.

\noindent\textbf{Theorem 1.} The automaton Q-learning algorithm given by
\begin{multline}
       Q_{t+1}(s_t, a_t) {=} (1 - \alpha_t)Q_t(s_t, a_t) +  \alpha_t \beta_t Q^\text{avg}_\text{teacher}(\omega, L(s_{t+1})) \\
       +  \alpha_t(1 - \beta_t)(R(s_t, a_t) + \gamma V_t(s_{t+1}))
    \end{multline}
converges to the optimal $Q^*(s,a)$ values if:
\begin{enumerate}
    \item The state and action spaces are finite. 
    \item $\alpha_t \in [0,1)$, $\sum_t \alpha_t = \infty$ and $\sum_t \alpha^2_t < \infty$.
    \item $\beta_t \geq 0$, $\lim\limits_{t \rightarrow \infty} \beta_t = 0$, and $\sum_t \alpha_t(1 - \beta_t) = \infty$.
    \item The variance $\operatorname{Var}(R(s,a))$ is bounded.
    \item $\gamma = 1$ and all policies lead to a cost-free terminal state; otherwise, $\gamma \in [0,1)$.
\end{enumerate}

\noindent\textbf{Proof:} We decompose automaton Q-learning into two parallel processes $q$ and $r$ given by

\begin{equation}
    \begin{aligned}
        q_{t+1}(s_t, a_t) & \hspace{-0.5mm}=\hspace{-0.5mm} (1 - \alpha_t)q_t(s_t, a_t) \\
        & +  \alpha_t(1 - \beta_t)(R(s_t, a_t) \hspace{-0.5mm}+ \hspace{-0.5mm} \gamma V_t(s_{t+1})) \\
        r_{t+1}(s_t, a_t) & \hspace{-0.5mm}=\hspace{-0.5mm} (1 - \alpha_t)r_t(s_t, a_t) \hspace{-0.5mm} + \hspace{-0.5mm} \alpha_t \beta_t Q^\text{avg}_\text{teacher}(\omega, L(s_{t+1}))
    \end{aligned}
\end{equation}
such that $Q_t(s,a) = q_t(s,a) + r_t(s,a)$. The process $q$ corresponds to a modified version of Q-update; using the same annealing function as in Equation~\eqref{equation:newLoss}, $q$ converges to the optimal Q-table $Q^*$ with probability (w.p.) 1 as shown in~\cite{jaakkola1993convergence}. To see that $r$ converges to 0 w.p. 1, we observe that $Q^\text{avg}_\text{teacher}(\omega, L_{s_{t+1}})$ is constant when training the student and so $r_t$ is a contraction map whose fixed point occurs at
\begin{equation}
    r_t(s_t, a_t) = \beta_t Q^\text{avg}_\text{teacher}(\omega, L(s_{t+1})).
\end{equation}
Since $\lim\limits_{t \rightarrow \infty} \beta = 0$, the fixed point of $r_t$ approaches 0 as $t \rightarrow \infty$. Thus, since $q$ and $r$ converge to $Q^*$ and $0$ respectively w.p. 1, their sum $Q$ converges to $Q^*$ w.p. 1. \qed \\

\noindent\textbf{Static Transfer.} Static value estimates can be effective when the abstract MDP defined by the automaton accurately captures the environment dynamics. Such estimates can be computed via tabular Q-learning over the abstract MDP:
\begin{align}
        Q(\omega, \sigma) \! \leftarrow \! Q(\omega, \sigma) & + \alpha (R(\omega, \sigma) \nonumber\\
        & + \gamma \max_{\sigma'} Q(\omega', \sigma') - Q(\omega, \sigma))\:.
    \label{equation:q}
\end{align}
The resulting Q-values can be used in the place of $Q^{\text{avg}}_{\text{teacher}}$ in Equation~\ref{equation:newLoss}. This method has the benefit of stabilizing training in the early stages without requiring a DQN oracle or any additional information beyond the reward structure.

\section{Related Work}
\label{sec:relwork}
Deep RL has made remarkable progress in many practical problems, such as recommendation systems, robotics, and autonomous driving. However, despite these successes, insufficient data and poor generalization remain open problems. In most real-world problems, it is difficult to obtain training data, so RL agents often learn with simulated data. However, RL agents trained with simulated data usually have poor performance when transferred to unknown environment dynamics in real-world data. To address these two challenges, transfer learning techniques~\cite{zhu2020transfer} have been adopted to solve two RL tasks: 1) state representation transfer and 2) policy transfer. 

Among transfer learning techniques, Domain Adaptation (DA) is the most well-studied in deep RL. Early attempts in DA constructed a map from states and actions in the source domain onto the target domain by hand~\cite{taylor2005behavior}. Subsequent efforts aimed to learn a set of general latent environment representations that can be transferred across similar domains but different state features. For example, in traditional policy distillation, knowledge is directly transferred from the teacher to the student \cite{rusu2015policy}. Also,
a multi-stage agent, DisentAngled Representation Learning Agent (DARLA)~\cite{higgins2017darla}, proposed to learn a general representation by adding an internal layer of pre-trained denoising autoencoder. With learned general representation in the source domain, the agent quickly learns a robust policy that produces decent performance on similar target domains without further tuning. However, DARLA cannot clearly define and separate domain-specific and domain-general features, and it causes performance degradation on some target tasks~\cite{higgins2017darla}.

Moreover, Contrastive Unsupervised Representations for Reinforcement Learning (CURL)~\cite{srinivas2020curl} aimed to address this issue by integrating contrastive loss. Furthermore, Latent Unified State Representation~\cite{xing2021domain} proposed a two-stage agent that can fully separate domain-general and domain-specific features by embedding both the forward loss and the reverse loss in Cycle-Consistent AutoEncoder~\cite{jha2018disentangling}. A critical difference between our approach and the aforementioned methods is its applicability when the state/action spaces of the student and teacher are different; for example, a discrete source and a continuous target environment as we demonstrate in the experiments.

Previous work has explored the use of high-level symbolic domain descriptions to construct a low-dimensional abstraction of the original state space~\cite{kokel2022hybrid}, which can be used to model the dynamics of the original system. Also, \cite{icarte2022reward} further construct an automaton which realizes the abstract decision process and used the automaton to convey information about the reward signal.

Static transfer learning methods based on reward machines were developed in~\cite{camacho2018non,icarte2022reward}. By treating the nodes and edges in the automaton as states and actions, respectively, the automaton can be transformed into a low-dimensional abstract MDP which can be solved using Q-learning or value iteration approaches.
The solution to the abstract MDP can speed up learning in the original environment using a potential-based reward shaping function~\cite{camacho2019ltl} or by introducing counterfactual experiences during training~\cite{icarte2022reward}.

\begin{figure}[t]
    \centering
    \includegraphics[width=\columnwidth]{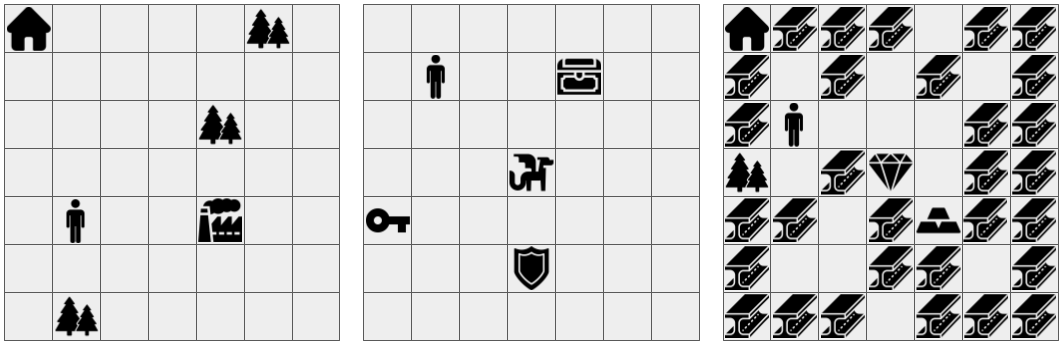} \\
    (a) \hspace{23mm} (b) \hspace{23mm} (c)
    \caption{Example $7 \times 7$ environment configurations for the \textit{Blind Craftsman} (a), \textit{Dungeon Quest} (b), and \textit{Diamond Mine} (c) environments.}
    \label{fig:envs}
\end{figure}

\begin{figure}[ht]
    \centering
    \includegraphics[width=6cm]{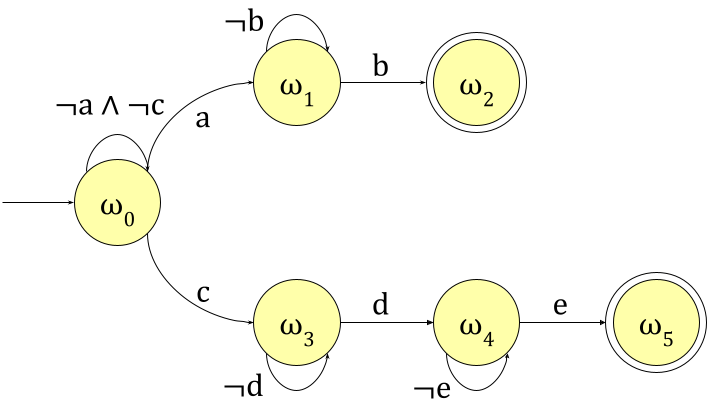}
    \caption{Simple automaton with two traces.}
    \label{fig:two_trace_aut}
\end{figure}

However, static transfer methods perform poorly when the abstract MDP fails to capture the behavior of the underlying process. Consider applying the static transfer approach proposed in~\cite{icarte2022reward} for the objective defined by the LTL$_\text{f}$ formula $\phi = \textbf{F}(\text{b} \vee \text{e}) \wedge (\neg \textbf{F}(\text{a}) \vee \neg \textbf{F}(\text{c})) \wedge (\text{a} \textbf{ R } \neg \text{b}) \wedge (\text{c} \textbf{ R } \neg \text{d}) \wedge (\text{d} \textbf{ R } \neg \text{e})$, whose automaton is given in Figure~\ref{fig:two_trace_aut}.

Assume that the reward function grants a reward of 1 for transitions leading to either terminal state and a reward of 0 for all other transitions.~\cite{icarte2022reward} perform value iteration over the abstract MDP with update
\begin{equation}
    V(\omega) := \max_{\omega' = \delta(\omega,\sigma)} R(\omega, \sigma) + \gamma V(\omega').
    \label{equation:value_iter}
\end{equation}
As can be seen in the automaton in Figure~\ref{fig:two_trace_aut}, there are two traces which satisfy the objective: one of length 2 and one of length 3. Due to the discount factor $\gamma < 1$, value iteration will favor taking transition $a$, which has a shorter accepting path, over transition $c$ in the starting state. However, it may be the case that observing $b$ after observing $a$ takes many steps in the original environment, and thus the longer trace $c \rightarrow d \rightarrow e$ takes less steps to reach an accepting state.

In contrast to static transfer, which uses prior knowledge of the reward function to model the behavior of the target process, dynamic transfer leverages experience acquired by interaction in a related domain to empirically estimate the target value function. Dynamic transfer has the advantage of implicitly factoring in knowledge of the teacher environment dynamics; in the previous example, if the shorter trace $a \rightarrow b$ takes more steps in the teacher environment than the longer trace $c \rightarrow d \rightarrow e$, this will be reflected in the discounted value estimates learned by the teacher.

\section{Experimental Results}
\label{sec:exp}
In this section, we evaluate our automaton distillation approach. For each time step in the student's training, we update our target policy with respect to $Q'$, following Algorithm \ref{alg:algorithm_automaton}.

\subsection{Evaluation}

We evaluate our algorithm on both grid-world and continuous environments. For the discrete environment, we use a randomly generated $7 \times 7$ instance of each grid-world environment comprising $49$ states as the source domain in which the teacher agent is trained. For the environment in which the student learns (the target domain) we use an independently generated $10 \times 10$ grid-world in the discrete-to-discrete knowledge transfer case. For the discrete-to-continuous knowledge transfer, we use a randomly generated an environment that is $7 \text{cm}$ long and $7 \text{cm}$ wide as the target domain in which the student learns. 
This continuous environment is illustrated in 
Figure \ref{fig:blind_craftsman_obstacles}(c), where objects are placed 
at random locations in the continuous space. For the grid-world, objects are placed arbitrarily on grid tiles and agents can obtain objects by being on the object tile. On the other hand, objects are collected in a continuous environment when their Euclidean distance to the object is $<0.4\,\text{cm}$. The environments we use are described below (sample $7 \times 7$ configurations for each environment can be seen in Figure~\ref{fig:envs}).

\textit{Blind Craftsman}:
This environment consists of woods, a factory, and a home; obstacles can be added for extra difficulty. The objective is satisfied when the agent has crafted three tools and arrived home. One wood is required to craft a tool. However,
since the agent can only carry two pieces of wood at a time, the agent must alternate between collecting wood and crafting tools. 

The objective is defined over the atomic propositions $AP = \{\text{wood}, \text{factory}, \text{tools} \geq 3, \text{home}\}$ and given by the LTL$_f$ formula $\phi = \textbf{G} (\text{wood} \implies \textbf{F} \text{ factory}) \wedge \textbf{F} (\text{tools} \geq 3 \text{ } \wedge \text{ home})$ with a corresponding automaton of 4 nodes and 12 transitions. 

\textit{Dungeon Quest}: This environment consists of a key, a chest, a shield, and a dragon. The agent can acquire a key and a shield by interacting with a key or shield, respectively. Additionally, the agent can obtain a sword by interacting with a chest with a key in its inventory. 

Once the agent has the sword and the shield, it may interact with the dragon to defeat it and complete the objective. 

The objective is defined over the atomic propositions $AP = \{\text{key}, \text{shield}, \text{sword}, \text{dragon}\}$ and given by the LTL$_f$ formula $\phi = \textbf{F}(\text{dragon}) \wedge (\text{key } \textbf{R } \neg \text{sword}) \wedge (\text{sword } \textbf{R } \neg \text{dragon}) \wedge (\text{shield } \textbf{R } \neg \text{dragon})$ with a corresponding automaton of 7 nodes and 17 transitions. 

\textit{Diamond Mine}: This environment consists of a wood tile, a diamond tile, gold tiles, and iron tiles.

\begin{figure*}[ht]
    
    \centering
    \includegraphics[width=\columnwidth]{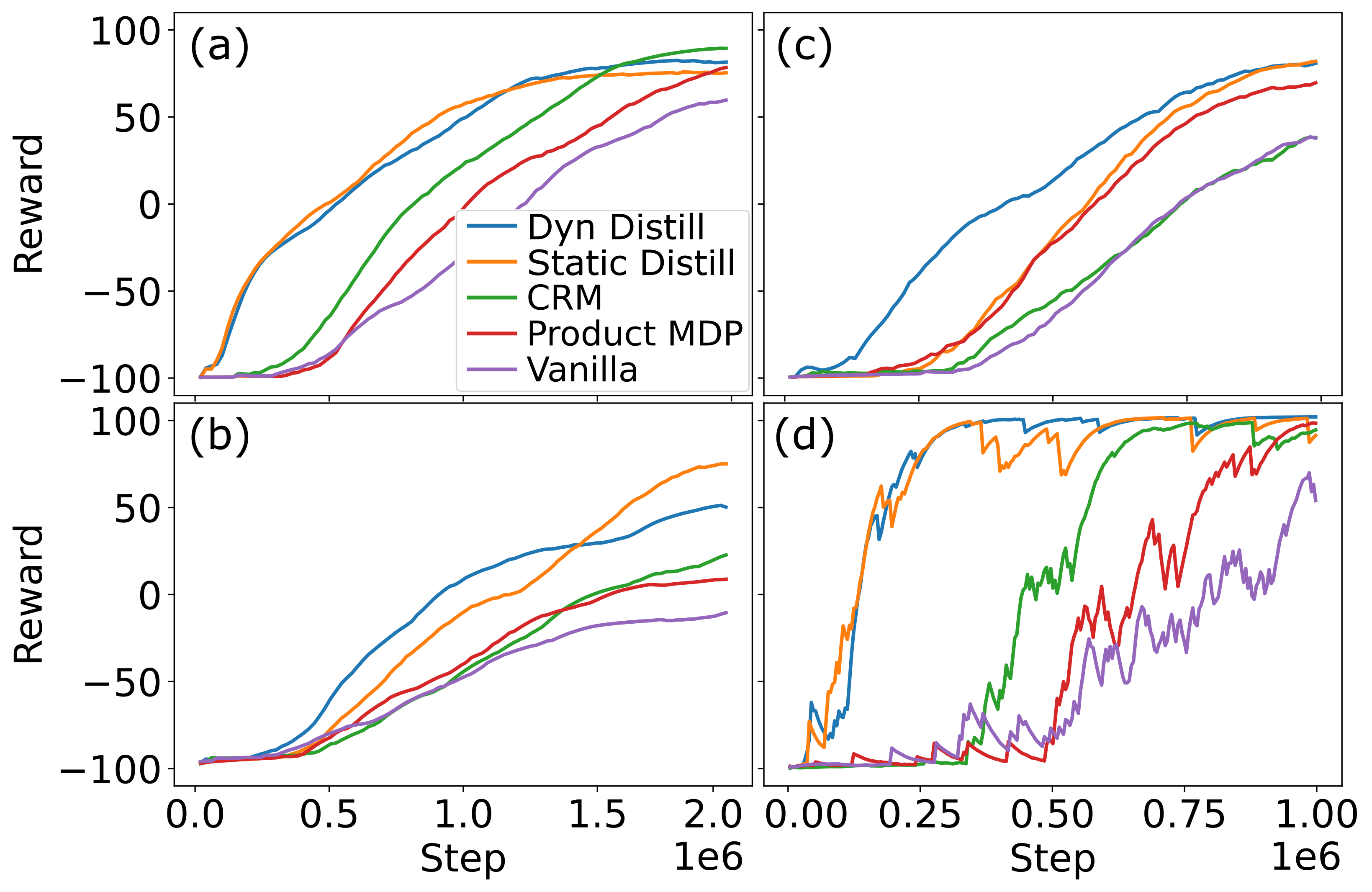} \includegraphics[width=\columnwidth,height=5.8cm]{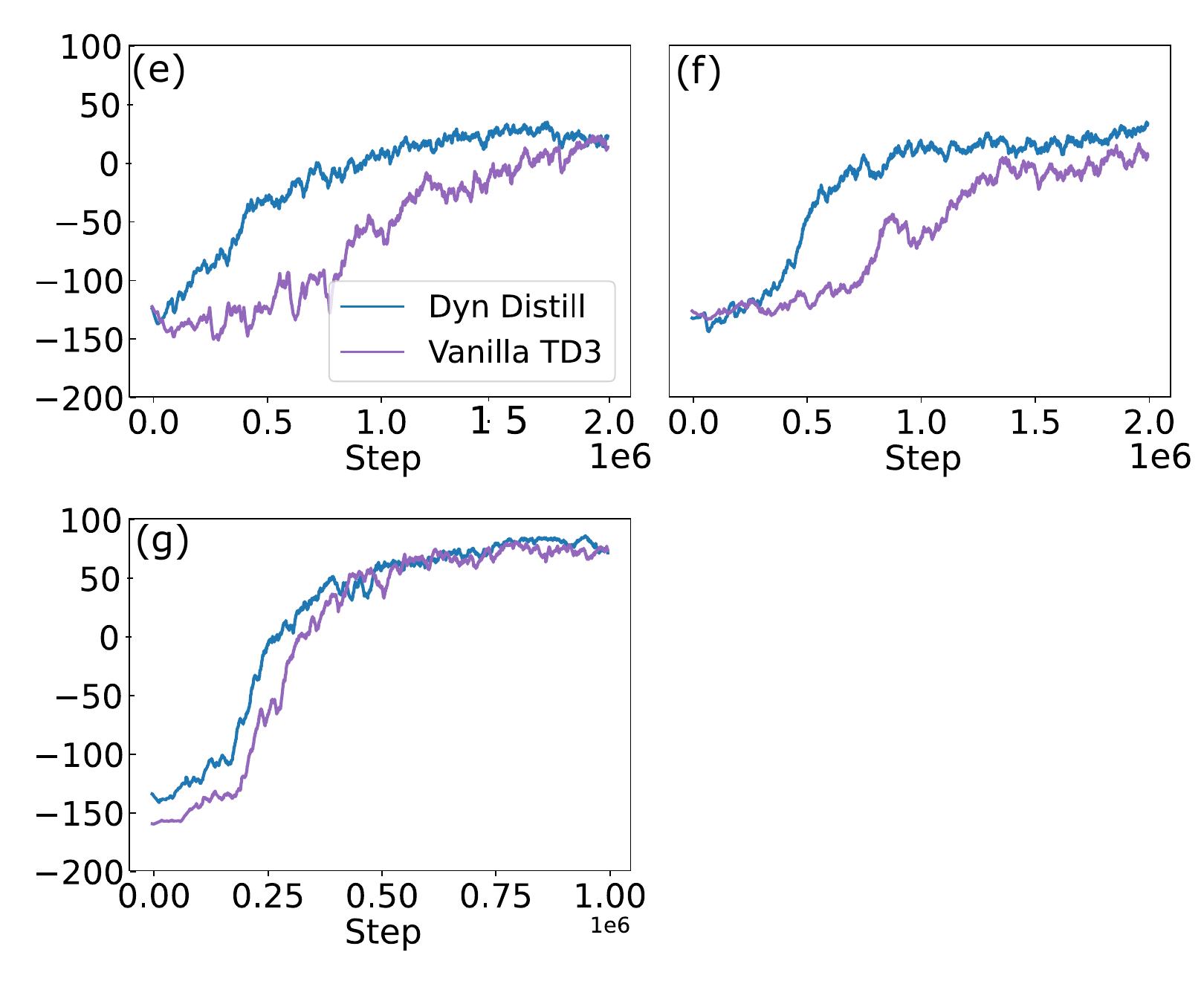}
    \caption{Reward per episode (y-axis) over time (x-axis) during training using dynamic automaton distillation (blue) vs. static automaton distillation (orange), Counterfactual Reward Machine (CRM) (green), Q-learning over the product MDP (red), and vanilla Q-learning (purple) on the \textit{Dungeon Quest} (a), {\textit{Diamond Mine}} (b) and \textit{Blind Craftsman} (c) environments. Each line is an ensemble average over 20 trials.
    \textit{Blind Craftsman with Obstacles} (see Figure~\ref{fig:blind_craftsman_obstacles}) is shown in (d). Dynamic automaton distillation for discrete to continuous  Blind Craftsman, Diamond Mine, and Dungeon Quest are captured in (e), (f), and (g), respectively.}
    \label{fig:training}
\end{figure*}
The agent may acquire wood, iron, or gold by interacting with the respective objects. Once the agent has collected wood and 30 iron, it automatically crafts a pickaxe. The agent may then obtain diamond by interacting with the diamond while holding a pickaxe. Once the agent has acquired either 1 diamond or 10 gold, it may return to home to complete the objective. To simplify the resulting automaton and limit unnecessary reward, once the agent has collected gold, it cannot obtain the diamond, and vice versa. Intuitively, although collecting diamond requires less automaton transitions, collecting 30 iron is relatively time-consuming. Thus, this environment represents a scenario where the objective automaton misrepresents the difficulty of the task. 

The objective is defined over the atomic propositions $AP = \{\text{wood}, \text{diamond}, \text{gold = 1}, \text{gold = 2}, ..., \text{gold = 10}, \text{home}\}$ and given by the LTL$_f$ formula $\phi = \textbf{F}(\text{home}) \wedge (\neg \textbf{F}(\text{gold} = 1) \vee \neg \textbf{F}(\text{wood})) \wedge (\text{wood} \textbf{ R } \neg \text{diamond}) \wedge (\text{gold = 1} \textbf{ R } \neg \text{gold = 2}) \wedge ... \wedge (\text{gold = 9} \textbf{ R } \neg \text{gold = 10}) \wedge ((\text{diamond} \vee \text{gold = 10}) \textbf{ R } \neg \text{home})$ with a corresponding automaton of 15 nodes and 29 transitions.

In the environments, the agent receives a reward of +1 for collecting each item, +100 for completing the task, and a -0.1 per time step, an additional -0.1 for going out of the boundary in the continuous environment. Agents are trained for 2 million play steps or until convergence. Each agent is represented by a Dueling DQN~\cite{wang2016dueling} or TD3~\cite{fujimoto2018addressing}.

\begin{figure*}[t]
    \centering
    \includegraphics[width=\textwidth]
    {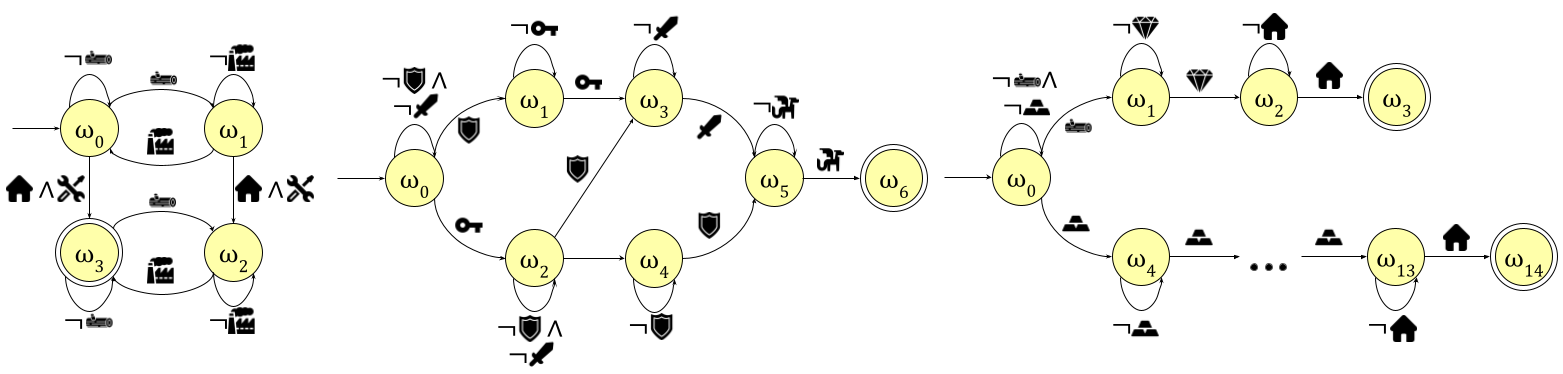} 
    \caption{Objective automata for the \textit{Blind Craftsman} (left), \textit{Dungeon Quest} (middle), and \textit{Diamond Mine} (right) environments.}
    \label{fig:automata}
\end{figure*}

The objective automaton is generated using the Python FLLOAT synthesis tool based on the LTL$_f$ behavioral specification; automata for each environment are shown in Figure~\ref{fig:automata}.
The automaton state is then tracked throughout each training episode and stored alongside each experience in the replay buffer. The automaton state is converted to a one-hot vector representation and concatenated to state representation.

In the continuous student environment, stability in the use of teacher's knowledge is implemented by delaying the actor-network policy updates every $d=4$ iterations for the first $20,000$ time steps. Afterwards, we use a more frequent actor update with $d=2$. This mimics the human learning process, where there is an initial understanding of foundational concepts that is applicable in different scenarios. 

In the discrete student, we evaluate the performance of static and dynamic automaton distillation against vanilla Q-learning in the student environment, Q-learning over the product MDP (i.e. including the automaton state in the DQN input), counterfactual experiences for reward machines (CRM), a static transfer method proposed in~\cite{icarte2022reward}. We observe that automaton distillation in a   $10\times10$ grid-world student environment outperforms the baselines (Figures \ref{fig:training}(a)-(c)) in the three environments, similarly, a faster convergence is seen in a student grid-world \textit{Blind Craftsman} environment with obstacles (fig. \ref{fig:training}(d)). 

For the student in a continuous environment, the results show a substantially faster convergence when compared to learning from scratch using TD3 (Fig. \ref{fig:training}(e)-(g)). Knowledge transfer methods such as policy distillation and DARLA are not compared against our proposed method as such methods are not structured to handle source and target environments with different structures. Specifically, in policy distillation~\cite{rusu2015policy}, the student uses the teacher's dataset to replicate the teacher's actions in a given state in a supervised learning manner, which assumes similarity in action and state spaces. 
Furthermore, DARLA and CURL focus on state representation learning, hence requiring similarities in the action space as well ~\cite{higgins2017darla,srinivas2020curl}. These limitations are not encountered using the Automaton for distillation.

Dynamic automaton distillation utilizes an empirical estimate of trajectory length over the teacher's decision process and circumvents inaccuracies in the abstract MDP. Thus, dynamic automaton distillation is effective when the optimal policies in the teacher and student environments follow similar automaton traces.
\begin{figure}
     \centering
     \includegraphics[width=0.5\columnwidth]{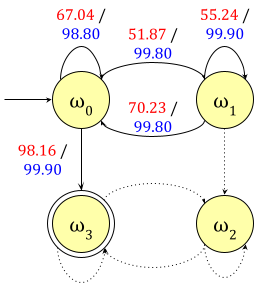} 
     \caption{Teacher Q-values produced by dynamic (red) and static (blue) automaton distillation on the \textit{Blind Craftsman} environment. Transitions which were not observed in the environment are denoted by dotted lines.}
     \label{fig:blind_craftsman_q}
\end{figure}
Some behavioral specifications can lead to objective automata with cycles, as evidenced by the \textit{Blind Craftsman} environment. Figure~\ref{fig:blind_craftsman_q} shows the Q-values generated by automaton distillation. Unlike static distillation, dynamic distillation can distinguish between states which are (nearly) equivalent in the abstract MDP by incorporating episode length; states which are further from the goal receive discounted rewards, resulting in smaller Q-values.

Cycles in the automaton do not necessarily lead to infinite reward loops as it is often the case that in the original environment, the cycle may be taken only a finite number of times. While it is possible to construct an automaton without cycles by expanding the state space of the automaton to include the number of cycles taken, the maximum number of cycle traversals must be known \textit{a priori} and incorporated into the objective specification, which may not be possible. Additionally, environments that share an objective may admit different numbers of cycle traversals; thus, cycles offer a compact representation that permits knowledge transfer between environments. However, cycles can aggravate the differences between the abstract MDP and the original decision process, resulting in negative knowledge transfer. In such cases, state-of-the-art transfer methods~\cite{icarte2022reward} may actually \textit{increase} training time relative to a na\"ive learning algorithm.

\section{Conclusion}
\label{sec:conc}
In this paper, we proposed automaton distillation, which leverages symbolic knowledge of the objective and reward structure in the form of formal language, to stabilize and expedite training of reinforcement learning agents. 

Value estimates for transitions in the automaton are generated using static (i.e. \textit{a priori}) methods such as value iteration over an abstraction of the target domain or dynamically estimated by mapping experiences collected in a related source domain to automaton transitions. The resulting value estimates are used as initial learning targets to bootstrap the student learning process resulting in faster learning even for a target environment different in structure from the source environment.
We illustrate several failure cases of existing automaton-based transfer methods, which exclusively reason over \textit{a priori} knowledge, and argue instead for the use of dynamic transfer. We demonstrate that both static and dynamic automaton distillation reduce training costs and outperform state-of-the-art knowledge transfer techniques.

\bibliography{reference}

\begin{thebibliography}{10}

\bibitem{tran2016deep}
Son~N Tran and Artur S~d’Avila Garcez.
\newblock Deep logic networks: Inserting and extracting knowledge from deep belief networks.
\newblock {\em IEEE transactions on neural networks and learning systems}, 29(2):246--258, 2016.

\bibitem{verma2018programmatically}
Abhinav Verma, Vijayaraghavan Murali, Rishabh Singh, Pushmeet Kohli, and Swarat Chaudhuri.
\newblock Programmatically interpretable reinforcement learning.
\newblock In {\em International Conference on Machine Learning}, pages 5045--5054. PMLR, 2018.

\bibitem{anderson2020neurosymbolic}
Greg Anderson, Abhinav Verma, Isil Dillig, and Swarat Chaudhuri.
\newblock Neurosymbolic reinforcement learning with formally verified exploration.
\newblock {\em Advances in neural information processing systems}, 33:6172--6183, 2020.

\bibitem{brunello2019synthesis}
Andrea Brunello, Angelo Montanari, and Mark Reynolds.
\newblock Synthesis of ltl formulas from natural language texts: State of the art and research directions.
\newblock In {\em 26th International Symposium on Temporal Representation and Reasoning (TIME 2019)}. Schloss Dagstuhl-Leibniz-Zentrum fuer Informatik, 2019.

\bibitem{wolper1983reasoning}
Pierre Wolper, Moshe~Y Vardi, and A~Prasad Sistla.
\newblock Reasoning about infinite computation paths.
\newblock In {\em 24th Annual Symposium on Foundations of Computer Science (sfcs 1983)}, pages 185--194. IEEE, 1983.

\bibitem{hasanbeig2021deepsynth}
Mohammadhosein Hasanbeig, Natasha~Yogananda Jeppu, Alessandro Abate, Tom Melham, and Daniel Kroening.
\newblock Deepsynth: Automata synthesis for automatic task segmentation in deep reinforcement learning.
\newblock In {\em Proceedings of the AAAI Conference on Artificial Intelligence}, volume~35, pages 7647--7656, 2021.

\bibitem{camacho2019ltl}
Alberto Camacho, Rodrigo~Toro Icarte, Toryn~Q Klassen, Richard~Anthony Valenzano, and Sheila~A McIlraith.
\newblock Ltl and beyond: Formal languages for reward function specification in reinforcement learning.
\newblock In {\em IJCAI}, volume~19, pages 6065--6073, 2019.

\bibitem{velasquez2021dynamic}
Alvaro Velasquez, Brett Bissey, Lior Barak, Andre Beckus, Ismail Alkhouri, Daniel Melcer, and George Atia.
\newblock Dynamic automaton-guided reward shaping for monte carlo tree search.
\newblock In {\em Proceedings of the AAAI Conference on Artificial Intelligence}, volume~35, pages 12015--12023, 2021.

\bibitem{de2015synthesis}
Giuseppe De~Giacomo and Moshe Vardi.
\newblock Synthesis for ltl and ldl on finite traces.
\newblock In {\em Twenty-Fourth International Joint Conference on Artificial Intelligence}, 2015.

\bibitem{zhu2017symbolic}
Shufang Zhu, Lucas~M Tabajara, Jianwen Li, Geguang Pu, and Moshe~Y Vardi.
\newblock Symbolic ltlf synthesis.
\newblock {\em arXiv preprint arXiv:1705.08426}, 2017.

\bibitem{taylor2005behavior}
Matthew~E Taylor and Peter Stone.
\newblock Behavior transfer for value-function-based reinforcement learning.
\newblock In {\em Proceedings of the fourth international joint conference on Autonomous agents and multiagent systems}, pages 53--59, 2005.

\bibitem{ammar2015autonomous}
Haitham~Bou Ammar, Eric Eaton, Jos{\'e}~Marcio Luna, and Paul Ruvolo.
\newblock Autonomous cross-domain knowledge transfer in lifelong policy gradient reinforcement learning.
\newblock In {\em Twenty-fourth international joint conference on artificial intelligence}, 2015.

\bibitem{wang2016dueling}
Ziyu Wang, Tom Schaul, Matteo Hessel, Hado Hasselt, Marc Lanctot, and Nando Freitas.
\newblock Dueling network architectures for deep reinforcement learning.
\newblock In {\em International conference on machine learning}, pages 1995--2003. PMLR, 2016.

\bibitem{van2016deep}
Hado Van~Hasselt, Arthur Guez, and David Silver.
\newblock Deep reinforcement learning with double q-learning.
\newblock In {\em Proceedings of the AAAI conference on artificial intelligence}, volume~30, 2016.

\bibitem{schaul2015prioritized}
Tom Schaul, John Quan, Ioannis Antonoglou, and David Silver.
\newblock Prioritized experience replay.
\newblock {\em arXiv preprint arXiv:1511.05952}, 2015.

\bibitem{lillicrap2015continuous}
TP~Lillicrap.
\newblock Continuous control with deep reinforcement learning.
\newblock {\em arXiv preprint arXiv:1509.02971}, 2015.

\bibitem{fujimoto2018addressing}
Scott Fujimoto, Herke Hoof, and David Meger.
\newblock Addressing function approximation error in actor-critic methods.
\newblock In {\em International conference on machine learning}, pages 1587--1596. PMLR, 2018.

\bibitem{jaakkola1993convergence}
Tommi Jaakkola, Michael Jordan, and Satinder Singh.
\newblock Convergence of stochastic iterative dynamic programming algorithms.
\newblock {\em Advances in neural information processing systems}, 6, 1993.

\bibitem{zhu2020transfer}
Zhuangdi Zhu, Kaixiang Lin, and Jiayu Zhou.
\newblock Transfer learning in deep reinforcement learning: A survey.
\newblock {\em arXiv preprint arXiv:2009.07888}, 2020.

\bibitem{rusu2015policy}
Andrei~A Rusu, Sergio~Gomez Colmenarejo, Caglar Gulcehre, Guillaume Desjardins, James Kirkpatrick, Razvan Pascanu, Volodymyr Mnih, Koray Kavukcuoglu, and Raia Hadsell.
\newblock Policy distillation.
\newblock {\em arXiv preprint arXiv:1511.06295}, 2015.

\bibitem{higgins2017darla}
Irina Higgins, Arka Pal, Andrei Rusu, Loic Matthey, Christopher Burgess, Alexander Pritzel, Matthew Botvinick, Charles Blundell, and Alexander Lerchner.
\newblock {DARLA}: Improving zero-shot transfer in reinforcement learning.
\newblock In {\em International Conference on Machine Learning}, pages 1480--1490. PMLR, 2017.

\bibitem{srinivas2020curl}
Aravind Srinivas, Michael Laskin, and Pieter Abbeel.
\newblock Curl: Contrastive unsupervised representations for reinforcement learning.
\newblock {\em arXiv preprint arXiv:2004.04136}, 2020.

\bibitem{xing2021domain}
Jinwei Xing, Takashi Nagata, Kexin Chen, Xinyun Zou, Emre Neftci, and Jeffrey~L Krichmar.
\newblock Domain adaptation in reinforcement learning via latent unified state representation.
\newblock In {\em Proceedings of the AAAI Conference on Artificial Intelligence}, volume~35, pages 10452--10459, 2021.

\bibitem{jha2018disentangling}
Ananya~Harsh Jha, Saket Anand, Maneesh Singh, and VS~Rao Veeravasarapu.
\newblock Disentangling factors of variation with cycle-consistent variational auto-encoders.
\newblock In {\em Proceedings of the European Conference on Computer Vision (ECCV)}, pages 805--820, 2018.

\bibitem{kokel2022hybrid}
Harsha Kokel, Nikhilesh Prabhakar, Balaraman Ravindran, Erik Blasch, Prasad Tadepalli, and Sriraam Natarajan.
\newblock Hybrid deep reprel: Integrating relational planning and reinforcement learning for information fusion.
\newblock In {\em 2022 25th International Conference on Information Fusion (FUSION)}, pages 1--8. IEEE, 2022.

\bibitem{icarte2022reward}
Rodrigo~Toro Icarte, Toryn~Q Klassen, Richard Valenzano, and Sheila~A McIlraith.
\newblock Reward machines: Exploiting reward function structure in reinforcement learning.
\newblock {\em Journal of Artificial Intelligence Research}, 73:173--208, 2022.

\bibitem{camacho2018non}
Alberto Camacho, Oscar Chen, Scott Sanner, and Sheila~A McIlraith.
\newblock Non-markovian rewards expressed in ltl: Guiding search via reward shaping (extended version).
\newblock In {\em GoalsRL, a workshop collocated with ICML/IJCAI/AAMAS}, 2018.

\end{thebibliography}
\bibliographystyle{unsrt}

\end{document}